\documentclass[11pt]{article}

\usepackage[margin=1in]{geometry}
\usepackage{amsmath, amssymb, amsfonts}
\usepackage{booktabs}
\usepackage{graphicx}
\usepackage{hyperref}
\usepackage{url}

\title{SkillChain-Gym: A Benchmark for Reskilling-Aware Production-Inventory Control under Disruptions}
\author{
Carlos Eduardo Sanoja\\
Quanta Labs, LLC\\
Professor, FCEA, Universidad Monte\'avila\\
Edificio Lomas del Sol, Calle Humboldt, Lomas del Sol, Caracas, Venezuela\\
\href{mailto:csanoja@somosquanta.com}{csanoja@somosquanta.com}\\
\href{https://orcid.org/0009-0000-0339-7072}{ORCID: 0009-0000-0339-7072}
}
\date{\today}

\begin{document}
\maketitle

\begin{abstract}
Production planning increasingly has to treat workforce capability as a
decision variable: certifications lapse when skills are not maintained, new
products require skills the current workforce does not hold, and reskilling
competes for the same worker hours that production needs today. Existing
operations benchmarks treat labor as exogenous, while workforce-planning
models with skills and learning are rarely released as reusable testbeds. We
introduce SkillChain-Gym, a benchmark specification for reskilling-aware
production-inventory control: a single-site production-inventory environment
with stylized worker skill-state dynamics, in which continuous skill levels
with hard threshold certification and forgetting are part of the state, and
training is a capacity-consuming action constrained by the same per-worker
time budget as production. The specification includes seed-controlled
disruption scenarios (demand spikes, absenteeism, and announced or surprise
new-product introductions requiring a rare skill), three feasibility modes
with mandatory projection diagnostics, deterministic replay, and metrics
covering operations, resilience, capability growth, and training-access
distribution. We evaluate an exact-feasible baseline taxonomy ---
production-only, reactive adaptive, water-filling adaptive, and
static-insurance policies with budget variants --- over 60-shift horizons
with paired statistical tests. The results are regime-dependent rather than
a ranking: training-capable policies dominate the production-only baseline
everywhere, and under realistic forgetting maintenance training is necessary
even without disruptions; between training-capable classes, adaptive training
prevails when bottlenecks are visible in the forecast, while a lean static
cross-training plan --- a deliberately favorable comparator whose structure
encodes the relevant skill contingencies --- acts as strong insurance under
surprise shocks and absenteeism, with capacity slack and the forgetting rate
governing the boundary between these regimes. No policy class dominates
across regimes, motivating forecast-driven controllers that decide when to
buy skill insurance and when to react.
\end{abstract}

\section{Introduction}
\label{sec:introduction}

Manufacturing planning increasingly has to treat workforce capability as a
decision variable rather than a fixed resource. Product transitions introduce
skills that the current workforce does not hold, certifications lapse when
skills are not maintained, absenteeism removes exactly the workers who carry a
scarce qualification, and reskilling programs compete for the same worker
hours that production needs today \cite{wef_reskilling_2026}. Planning systems
that ignore this coupling can meet service targets right up to the moment a
skill bottleneck binds --- and then have no feasible action left, because
certifications cannot be bought instantaneously.

Two research communities study the ingredients of this problem, largely in
separation. On one side, operations and supply-chain benchmark environments
--- OR-Gym, MABIM, SafeOR-Gym, and related suites
\cite{Hubbs2020ORGym,Yang2023MABIM,Ramanujam2025SafeORGym,Berto2025RL4CO} ---
provide reusable, seed-controlled environments for inventory and production
control with standardized baselines, but they treat labor capacity as
exogenous or absent: no skill state, no training action, no forgetting. On
the other side, workforce-planning and dual-resource scheduling research has
long modeled skills, cross-training, learning, and forgetting
\cite{DeBruecker2015SkillsReview,SaidiMehrabad2013WorkerTraining,Hopp2004SkillChaining,Heuser2022BudgetedTraining},
but these models are typically bespoke optimization studies rather than
reusable environments with common interfaces, baselines, and metrics. As a
result, there is no standard testbed on which production-planning policies
that \emph{decide about training} can be compared.

This paper introduces \textbf{SkillChain-Gym}, a benchmark specification for
reskilling-aware production and inventory control: a single-site environment
with stylized worker skill-state dynamics.
Each worker carries a continuous skill level per skill; certification is a
hard threshold on that level; production requires certification; skills decay
unless maintained; and training is an explicit action that consumes the same
worker hours as production (Section~\ref{sec:formulation}). This last
property --- training as a \emph{capacity-consuming} action --- is the
benchmark's central mechanism. When training is free or instantaneous, skill
management decouples from operations and reduces to a scheduling detail.
When every training hour is a production hour forgone, policies face a
genuine intertemporal trade-off: spend capacity now to serve demand, or
invest it to remain able to serve demand later. Disruption scenarios
(demand spikes, absenteeism, and announced or surprise new-product
introductions that require a rare skill) stress exactly this trade-off.

Our empirical contribution is a \emph{regime map} rather than a winner. Across
nine T$=60$ scenario instances, 20--50 seeds per cell, and paired
statistical tests, we find that training-capable policies dominate a
production-only baseline everywhere --- under realistic forgetting,
maintenance training is mandatory even without shocks --- but that the
comparison \emph{between} training-capable policy classes is governed by
three parameters: bottleneck visibility, capacity slack, and the forgetting
rate. Forecast-visible bottlenecks favor adaptive, reactive training, which
beats every over-provisioned static plan we tested. Surprise shocks and
absenteeism favor a lean static cross-training plan, which acts as insurance
whose cost is the pre-shock labor it diverts; near the demand--capacity
boundary, where reaction transients become structurally unrecoverable, this
insurance dominates regardless of the forgetting rate. We deliberately report
the static plan as a \emph{favorable} comparator --- its structure encodes
which skills can become critical --- and we isolate, via a water-filling
allocation variant, how much of the adaptive policies' shortfall is an
allocation artifact rather than a policy-class property.

Concretely, our contributions are:
\begin{itemize}
  \item \textbf{A benchmark specification} for reskilling-aware
    production-inventory control: a single-site, Gymnasium-style environment
    in which worker capability is part of the state and training is part of
    the action space, with deterministic seeded replay and three feasibility
    modes with mandatory projection diagnostics
    (Section~\ref{sec:benchmark}).
  \item \textbf{A locked formulation} of stylized worker skill-state dynamics:
    continuous skill levels, hard threshold certification, capacity-consuming
    training, and forgetting, with soft productivity and learning-by-doing
    confined to labeled extensions (Section~\ref{sec:formulation}).
  \item \textbf{Disruption scenarios and metrics} covering operations,
    resilience (including recovery rates and unrecovered-episode counts),
    capability growth, and the distribution of training access across workers
    (Sections~\ref{sec:design-scenarios} and~\ref{sec:metrics}).
  \item \textbf{A baseline taxonomy} spanning production-only, reactive
    adaptive, water-filling adaptive, and static-insurance policies with
    budget variants, all exact-feasible and none tuned to win
    (Section~\ref{sec:baselines}).
  \item \textbf{A regime analysis} showing that no policy class dominates:
    adaptive training helps when bottlenecks are visible or forecasted, while
    lean static cross-training can act as strong insurance under surprise
    shocks and absenteeism, with capacity slack and forgetting as the
    governing parameters (Section~\ref{sec:results}).
\end{itemize}

These findings position the benchmark as a foundation rather than an
endpoint: the regime structure --- visibility, slack, forgetting --- is
precisely the setting in which receding-horizon controllers that value future
skill capacity should be able to combine the strengths of both policy
classes, which we study in a companion paper. The remainder of this paper
reviews related work (Section~\ref{sec:related}), formalizes the model
(Section~\ref{sec:formulation}), describes the benchmark design
(Section~\ref{sec:benchmark}), and presents the experimental setup, results,
and discussion (Sections~\ref{sec:experiments}--\ref{sec:discussion}).

\section{Related Work}
\label{sec:related}

\paragraph{Operations and supply-chain reinforcement-learning benchmarks.}
Several recent environments make operations research problems accessible to reinforcement learning through standardized benchmark interfaces. OR-Gym introduced Gym-style environments for classical OR problems including multi-echelon supply chains, with RL policies compared against optimization and heuristic baselines \cite{Hubbs2020ORGym}. MABIM extends this direction to a multi-agent, multi-echelon, multi-commodity inventory simulator for inventory-management research \cite{Yang2023MABIM}. Alvo et al. argue that inventory networks are a promising setting for reliable policy optimization and release benchmark environments for inventory-network control \cite{Alvo2023InventoryNetworks}. SafeOR-Gym further adapts practical OR environments to constrained Markov decision processes for safe reinforcement learning \cite{Ramanujam2025SafeORGym}, while RL4CO provides a broad benchmark framework for reinforcement learning in combinatorial optimization \cite{Berto2025RL4CO}. These benchmarks establish the value of reusable environments, baselines, and standardized evaluation, but they generally treat labor capacity as exogenous or absent rather than as a dynamic capability shaped by training decisions.

\paragraph{Reinforcement learning for inventory and supply-chain control.}
The supply-chain RL literature is active and diverse, as summarized by Rolf et al. \cite{Rolf2022SCMRLReview}. Prior work has studied constrained continuous-action RL for inventory management \cite{Burtea2024ConstrainedInventoryRL}, deep RL for multi-echelon inventory systems \cite{Geevers2024MultiEchelonDRL}, risk-sensitive and distributional RL for multi-echelon supply chains \cite{Wu2023InventoryManagementCAChE,Wu2023DistributionalInventory}, bandit-based inventory optimization \cite{Preil2022BanditInventory}, and disruption-aware multi-echelon inventory policies \cite{Lu2025DisruptiveInventoryPPO}. These works model inventories, replenishment, lead times, costs, and disruptions, but they do not make worker skills, learning, forgetting, or reskilling actions part of the environment state and action spaces.

\paragraph{Workforce planning, skills, training, and learning.}
Workforce planning with skills is a mature literature. De Bruecker et al. review technical and managerial aspects of skill-aware workforce planning, including skill classes, substitution, cross-training, and learning effects \cite{DeBruecker2015SkillsReview}. Production planning with worker training has also been modeled directly: Saidi-Mehrabad et al. formulate a dynamic manufacturing planning model that includes worker assignment, worker training, machine time, inventory, and backorder costs \cite{SaidiMehrabad2013WorkerTraining}. Heuser et al. study flexible and budgeted training under volatile demand with learning-by-doing and forgetting \cite{Heuser2022BudgetedTraining}; Valeva et al. analyze the trade-off between workforce flexibility and inventory when workers learn through experience under demand uncertainty \cite{Valeva2017FlexInventory}; and Cavagnini et al. model uncertain learning rates in workforce production planning \cite{Cavagnini2020UncertainLearning}. Ruf et al. formulate an MDP for hierarchical skills, long-term training, and random resignations in workforce capacity planning \cite{Ruf2022HierarchicalSkills}. These studies show that reskilling-aware production planning is not new as an optimization topic. The gap is that these models are typically not released as reusable benchmark environments with common RL/control baselines and standardized capability and training-access metrics.

\paragraph{Dual-resource scheduling.}
The dual-resource constrained job-shop literature shows that cross-training and worker flexibility can materially affect shop performance. Early simulation and analytical studies examined cross-training in job shops \cite{Park1991CrossTraining}, skill chaining in serial production lines \cite{Hopp2004SkillChaining}, worker transfer delays and learning losses \cite{Kher1994WorkerFlexibility}, and heterogeneous flexibility under learning and forgetting \cite{Yue2008WorkerFlexibility,Felan2001HeterogeneousFlexibility}. More recent work integrates learning and forgetting into worker assignment and production planning \cite{Liu2016WorkerAssignment}, job rotation with skill and motivation variation \cite{Azizi2010JobRotation}, product-category learning and forgetting \cite{Heuser2023SingleMachineLearningForgetting}, competence-aware mixed-model assembly scheduling \cite{Palao2023CompetenceAware,Palao2025ResilientWorkforce}, and robust technician training/resource planning for product evolution \cite{Karimi2024TechnicianResource}. Recent sociotechnical scheduling and human-centered production-planning systems also include worker capabilities, preferences, or fatigue \cite{Heide2026AIAssistance,Grumbach2023Sociotechnical,Hu2025GraphDRLWorkerFatigue}. SkillChain-Gym therefore does not claim to introduce worker skills into production scheduling; instead, it targets a benchmark gap at the intersection of production-inventory control, explicit reskilling actions, disruption scenarios, and human capability metrics.

\paragraph{Reliability and training-access metrics.}
Industrial reinforcement-learning systems require more than high average reward. Waubert de Puiseau et al. survey reliability concepts for RL-based production scheduling and emphasize robustness, stability, and risk-aware evaluation \cite{Waubert2022ReliabilitySurvey}. How training opportunities are distributed across workers also matters once they become algorithmically allocated. Formal work on fair reinforcement learning studies sequential settings where actions affect future states and rewards \cite{Jabbari2017FairnessRL}, while Weng frames fairness through social welfare functions in RL \cite{Weng2019FairnessRL}. For resource allocation, Jain's index provides a simple quantitative dispersion measure \cite{Jain1984FairnessIndex}, and Yilmaz et al. study fairness--efficiency trade-offs in collaborative demand and capacity sharing \cite{Yilmaz2017FairSharing}. SkillChain-Gym draws on these ideas at the level of \emph{training-access metrics}: it reports per-worker training hours, minimum access, and Jain/Gini dispersion alongside operational metrics, making visible policies that achieve short-term service by concentrating training on few workers. No demographic attributes are modeled, and no broader fairness claims are made.

\paragraph{Positioning.}
The reviewed literature suggests a narrow but meaningful contribution. Existing benchmarks cover inventory, scheduling, constrained OR, and combinatorial optimization; existing workforce models cover skill-aware assignment, learning, forgetting, and training. What is missing is a reusable benchmark specification that joins these lines: a single-site production-inventory environment with stylized worker skill-state dynamics, in which reskilling is an explicit capacity-consuming action, disruption scenarios create skill bottlenecks, and policies are evaluated with operational, resilience, capability, and training-access metrics under a shared API. Consistent with this positioning, our empirical findings are regime-dependent rather than a ranking: no baseline class dominates across visibility, capacity-slack, and forgetting regimes (Section~\ref{sec:results}).

\section{Problem Formulation}
\label{sec:formulation}

SkillChain-Gym is a finite-horizon episodic Markov decision process in which a
planner allocates scarce worker time between production and reskilling at a
single manufacturing site. One period corresponds to one work shift. The
formulation below is the default model; all extensions are explicitly labeled
as such and are excluded from the main results.

\subsection{Sets and parameters}
\label{sec:sets}

Let $t \in \{0,\dots,T-1\}$ index shifts, $p \in \mathcal{P}$ products,
$w \in \mathcal{W}$ workers, and $k \in \mathcal{K}$ skills. Each product
requires one primary skill $k(p) \in \mathcal{K}$; multi-skill products are an
extension. Static parameters are: productivity $v_p > 0$ (units per certified
worker-hour), nominal worker hours $H_w > 0$ per shift, certification
thresholds $\theta_k \in [0,1]$, training gain $\alpha^{\mathrm{train}}_{w,k}
\ge 0$ per training hour, forgetting rates $\delta_k \ge 0$ per shift,
training-seat capacities $cap^{\mathrm{train}}_{k,t}$, aggregate
production-capacity hours $C_t$, and cost coefficients
$c^{B}_p$ (backlog), $c^{I}_p$ (holding), $c^{q}_p$ (production), and
$c^{Y}_{w,k}$ (training) per unit and shift.

\subsection{State}
\label{sec:state}

The state at shift $t$ is
\begin{equation}
  x_t = \bigl(I_t,\, B_t,\, \hat{D}_t,\, A_t,\, C_t,\, S_t,\, Q_t,\, Z_t,\, \tau_t\bigr),
  \label{eq:state}
\end{equation}
where $I_{p,t} \ge 0$ is on-hand inventory, $B_{p,t} \ge 0$ is backlog,
$\hat{D}_{p,t:t+F}$ is the demand forecast over a window of $F$ shifts,
$A_{w,t} \in [0, H_w]$ is available worker time, $C_t \ge 0$ is the aggregate
production-capacity pool, $S_{w,k,t} \in [0,1]$ is the continuous skill level
of worker $w$ in skill $k$, $Z_t$ is the disruption state, and $\tau_t$
collects time features. Certification is \emph{derived} from the skill level
by hard thresholding:
\begin{equation}
  Q_{w,k,t} \;=\; \mathbf{1}\!\left[S_{w,k,t} \ge \theta_k\right] \in \{0,1\}.
  \label{eq:certification}
\end{equation}
There is no separate training-progress state in the default model, no soft
(graded) productivity, and no learning-by-doing; the only default mechanism
that raises $S$ is explicit training, and the only mechanism that lowers it is
forgetting.

\subsection{Actions}
\label{sec:actions}

The action is a pair of nonnegative worker-hour allocations,
\begin{equation}
  u_t = \bigl(a^{\mathrm{prod}}_t,\, a^{\mathrm{train}}_t\bigr), \qquad
  a^{\mathrm{prod}}_{w,p,t} \ge 0, \quad a^{\mathrm{train}}_{w,k,t} \ge 0,
  \label{eq:action}
\end{equation}
where $a^{\mathrm{prod}}_{w,p,t}$ assigns worker $w$'s hours to producing
product $p$ and $a^{\mathrm{train}}_{w,k,t}$ assigns them to training skill
$k$. Both draw on the \emph{same} per-worker time budget:
\begin{equation}
  \sum_{p} a^{\mathrm{prod}}_{w,p,t} \;+\; \sum_{k} a^{\mathrm{train}}_{w,k,t}
  \;\le\; A_{w,t} \qquad \forall w.
  \label{eq:budget}
\end{equation}
This opportunity-cost constraint is the central mechanism of the benchmark:
reskilling is capacity-consuming, so every training hour is a production hour
forgone.

\subsection{Feasibility}
\label{sec:feasibility}

Production eligibility is hard: uncertified workers cannot produce,
\begin{equation}
  a^{\mathrm{prod}}_{w,p,t} = 0 \quad \text{whenever} \quad Q_{w,k(p),t} = 0.
  \label{eq:eligibility}
\end{equation}
Aggregate production hours are capped by the capacity pool and training hours
by per-skill seats:
\begin{equation}
  \sum_{w,p} a^{\mathrm{prod}}_{w,p,t} \le C_t,
  \qquad
  \sum_{w} a^{\mathrm{train}}_{w,k,t} \le cap^{\mathrm{train}}_{k,t}
  \quad \forall k.
  \label{eq:capacity}
\end{equation}
Production output is
\begin{equation}
  q_{p,t} \;=\; v_p \sum_{w} a^{\mathrm{prod}}_{w,p,t}\, Q_{w,k(p),t},
  \label{eq:production}
\end{equation}
where the multiplication by $Q$ is a simulator safeguard: feasible actions
already satisfy \eqref{eq:eligibility}, and the environment reports the hours
it zeroes as a diagnostic (Section~\ref{sec:feasibility-modes}).

\subsection{Transition dynamics}
\label{sec:dynamics}

Demand $D_{p,t}$ is realized after the action as the forecast mean plus
seed-controlled noise, lower-censored at zero and gated to active products.
Shipments serve current demand plus backlog from inventory and current
production:
\begin{align}
  \mathrm{ship}_{p,t} &= \min\bigl(I_{p,t} + q_{p,t},\; D_{p,t} + B_{p,t}\bigr),
  \label{eq:ship}\\
  I_{p,t+1} &= I_{p,t} + q_{p,t} - \mathrm{ship}_{p,t},
  \label{eq:inventory}\\
  B_{p,t+1} &= B_{p,t} + D_{p,t} - \mathrm{ship}_{p,t},
  \label{eq:backlog}
\end{align}
so $I_{p,t} \ge 0$ and $B_{p,t} \ge 0$ by construction. Skills decay
geometrically and grow linearly in training time, clipped to $[0,1]$:
\begin{equation}
  S_{w,k,t+1} \;=\;
  \Pi_{[0,1]}\!\Bigl((1-\delta_k)\, S_{w,k,t}
  + \alpha^{\mathrm{train}}_{w,k}\, a^{\mathrm{train}}_{w,k,t}\Bigr),
  \qquad
  Q_{w,k,t+1} = \mathbf{1}\!\left[S_{w,k,t+1} \ge \theta_k\right].
  \label{eq:skill}
\end{equation}
A learning-by-doing term ($S$ increasing in production hours on the matching
skill) is supported only as an explicitly labeled extension and is not used in
any reported result.

\subsection{Objective}
\label{sec:objective}

The per-shift cost aggregates backlog, holding, production, and training
costs, plus a violation penalty $c^{\mathrm{viol}}_t$ that is zero for
feasible actions:
\begin{equation}
  c_t \;=\; \sum_{p} c^{B}_p B_{p,t+1}
  + \sum_{p} c^{I}_p I_{p,t+1}
  + \sum_{p} c^{q}_p q_{p,t}
  + \sum_{w,k} c^{Y}_{w,k}\, a^{\mathrm{train}}_{w,k,t}
  + c^{\mathrm{viol}}_t,
  \label{eq:cost}
\end{equation}
and the reward is $r_t = -c_t$. The scalar reward is used for ranking, but
all results additionally report decomposed operational, resilience,
capability, and training-access metrics
(Section~\ref{sec:metrics}); benchmark success is defined by exposing
meaningful trade-offs, not by any single policy maximizing $\sum_t r_t$.

\section{Benchmark Design}
\label{sec:benchmark}

SkillChain-Gym is a single-site production-inventory benchmark with stylized
worker skill-state dynamics. Its purpose is to expose, in a reusable
Gymnasium-style environment, the trade-off created when the same scarce worker
time must cover both production and capacity-consuming reskilling, under
forgetting and under disruptions that stress skill coverage. This section
describes the environment realization of the formulation in
Section~\ref{sec:formulation}; the default instance used throughout the
experiments has two products, three skills, four workers, and one aggregate
production-capacity pool. The interface conventions follow established
OR/RL benchmark environments
\cite{Hubbs2020ORGym,Yang2023MABIM,Ramanujam2025SafeORGym}, which treat labor
as exogenous; here worker capability is part of the state and action spaces.

\subsection{Environment and interface}
\label{sec:env-api}

The environment follows the standard episodic interface
(\texttt{reset} and \texttt{step}), subclassing Gymnasium when it is available
and exposing the same signatures without
the dependency otherwise. Observations expose inventory, backlog, the demand
forecast window, worker availability, the capacity pool, the skill matrix
$S$, the derived certification matrix $Q$ \eqref{eq:certification}, the
product--skill map, and the shift index. Actions are the worker-hour
matrices of \eqref{eq:action}. Episodes are fully seed-controlled: a fixed
seed reproduces demand realizations, scenario randomization, and therefore
every reported number bit-for-bit.

\subsection{Hard certification and capacity-consuming training}
\label{sec:design-skills}

Two locked semantics define the benchmark's character. First, certification
is hard \eqref{eq:certification}--\eqref{eq:eligibility}: a worker either is
or is not eligible to produce a product, with no partial productivity below
the threshold. Second, training consumes the same time budget as production
\eqref{eq:budget}, so skill acquisition is never free. Combined with
forgetting \eqref{eq:skill}, these semantics generate the phenomena the
experiments measure: certifications erode unless maintained, rare-skill
bottlenecks cannot be bought instantaneously, and insurance-style
cross-training competes with current output. The linear-gain/geometric-decay
abstraction is a deliberate simplification of the rich literature on workforce
training, learning, and forgetting in production settings
\cite{DeBruecker2015SkillsReview,Heuser2022BudgetedTraining,Yue2008WorkerFlexibility,Ruf2022HierarchicalSkills};
we make no claim of empirical calibration to real workforce behavior.

\subsection{Feasibility modes and projection diagnostics}
\label{sec:feasibility-modes}

The environment supports three feasibility modes. In \texttt{project} (the
default), infeasible continuous actions are deterministically repaired:
negative entries are clipped, ineligible production hours are zeroed, and
per-worker budgets, the capacity pool, and training seats are rescaled to
feasibility. In \texttt{strict}, the repaired action is executed but
pre-projection violations incur a cost penalty. In \texttt{masked}, the
environment exposes a per-worker eligibility-and-budget mask and raises an
error on any infeasible action, with no silent repair. Because projection can
distort policy comparisons, every table reports projection frequency, mean
projection norm, per-constraint violation counts, and
certification-zeroed production hours; all baselines in this paper emit
exactly feasible actions, so these diagnostics are identically zero in the
reported runs and projection never influences the results.

\subsection{Scenario families}
\label{sec:design-scenarios}

Episodes are generated from four seed-controlled scenario families, each
specifying shock start, duration, magnitude, affected entities, and forecast
visibility: (i) \emph{no-shock sanity}; (ii) \emph{demand shock}, a temporary
spike for one product; (iii) \emph{absenteeism shock}, removing selected
workers' availability for a window; and (iv) \emph{new-product skill shock},
activating a product whose primary skill no worker initially holds, in an
\emph{announced} variant (visible in the forecast window before activation)
and a \emph{surprise} variant (activation shift randomized per seed and
hidden from the forecast until onset). The experimental instances,
shock-window placement, and demand/capacity scalings are detailed in
Section~\ref{sec:scenarios}.

\subsection{Metrics}
\label{sec:metrics}

Beyond scalar cost \eqref{eq:cost}, the benchmark reports four metric groups:
\emph{operational} (total cost, service level, per-product service, total and
peak backlog, throughput, utilization); \emph{resilience} (recovery time
relative to the pre-shock backlog level, recovery rate, unrecovered-episode
counts, worst-shift service); \emph{capability} (average skill gain, new
certifications, skill-bottleneck severity); and \emph{training-access}
(training hours per worker, minimum access, Jain index, Gini coefficient).
Training-access metrics are reported as observables of how policies
distribute training time across workers; no demographic attributes are
modeled and no fairness claims are attached to them.

\subsection{Baseline classes}
\label{sec:design-baselines}

The benchmark ships five baseline classes, all emitting exactly feasible
actions: a \emph{production-only} policy (no training; lower-bounds what is
achievable without skill investment); \emph{reactive adaptive} policies that
train toward the largest anticipated certified-capacity shortfall (a myopic
greedy allocator and a deliberately slow-reacting variant); a
\emph{water-filling adaptive} policy that differs from the myopic one only in
splitting production capacity proportionally to need, isolating allocation
artifacts from policy-class effects; a \emph{static insurance} policy that
executes a fixed open-loop cross-training plan in the first shifts and never
observes shocks; and \emph{static budget variants} that scale only the plan's
hours per shift. A feasible random policy anchors the tables. The static
plan's structure encodes which skills can become critical in this scenario
family, which makes it a deliberately favorable insurance comparator
(Section~\ref{sec:baselines}); rolling-horizon optimization and learned
policies are deferred to future work.

\subsection{Intentional scope restrictions}
\label{sec:scope}

The default model deliberately excludes procurement and material pipelines,
supplier delays, multi-echelon flows, detailed job-shop routing and
station-specific capacities, soft (graded) productivity, and
learning-by-doing. These omissions are design choices, not oversights: each
added mechanism would confound the question the benchmark isolates --- how
policies trade current production against future skill capacity under an
explicit opportunity-cost constraint. A benchmark in which material
availability, routing, or graded productivity also bind would make it
impossible to attribute outcome differences to skill decisions; the
single-site, single-pool, hard-certification instance keeps every baseline
interpretable and makes the regime analysis of
Section~\ref{sec:results} attributable to visibility, slack, and forgetting
alone. Learning-by-doing and soft productivity are supported by the
implementation as explicitly labeled extensions for ablation studies, and
richer scenario families (including instances with enough skills that no
static plan can pre-train every contingency) are left as future work.

\section{Experimental Setup}
\label{sec:experiments}

All experiments use the default benchmark instance of
Sections~\ref{sec:formulation} and~\ref{sec:benchmark} unless a sensitivity
axis is explicitly varied:
two products, three skills, four workers, one aggregate production-capacity
pool, hard threshold certification ($\theta_k = 0.6$;
Eq.~\ref{eq:certification}), explicit capacity-consuming training
($\alpha^{\mathrm{train}} = 0.05$ per hour; Eq.~\ref{eq:budget}), forgetting
($\delta_k = 0.005$ per shift; Eq.~\ref{eq:skill}), and a horizon of
$T = 60$ shifts.
Instances are synthetic and fully seed-controlled; every reported number is
reproducible bit-for-bit from a fixed seed, and each suite re-runs a subset of
episodes to assert deterministic replay. All policies are evaluated in the
\texttt{project} feasibility mode, but every baseline emits exactly feasible
actions by construction: projection frequency and projection norm are zero in
all reported runs, and we additionally report certification-zeroed production
hours as a diagnostic. No real data are used.

\subsection{Scenario families}
\label{sec:scenarios}

We evaluate nine scenario instances drawn from four families, with shock
windows placed early, mid, and late in the horizon (shifts 10, 26, and 44;
duration 8) so that resilience metrics are not diluted by a long quiet tail:

\begin{itemize}
  \item \textbf{No-shock sanity}: stationary demand, full availability.
  \item \textbf{Demand shock} (early/mid/late): a temporary demand spike for
    one product.
  \item \textbf{Absenteeism shock} (early/mid/late): the two workers holding
    the second skill are unavailable during the window.
  \item \textbf{New-product skill shock}: a product activates mid-episode and
    requires a rare skill for which no worker is initially certified. In the
    \emph{announced} variant the activation is visible in the demand forecast
    window before it occurs; in the \emph{surprise} variant the activation
    shift is randomized per seed (uniform over shifts 4--48) and hidden from
    the forecast until onset.
\end{itemize}

In the new-product scenarios, post-activation demand ($\approx 30$ labor-hours
per shift) deliberately sits near the capacity pool (32 hours) to stress
allocation under scarcity; we vary this slack explicitly in
Section~\ref{sec:results-boundary}. The total labor envelope (4 workers
$\times$ 8 hours) is a second binding resource; scenario scalings must respect
it, and we document one discarded design that did not.

\subsection{Baseline taxonomy}
\label{sec:baselines}

The baselines are organized into five classes; all are exact-feasible and none
is tuned to win on any particular scenario.

\begin{description}
  \item[Production-only.] \textsc{GreedyProduction} allocates all worker time
    to production with a scarcity-aware greedy rule and never trains. It lower
    bounds what is achievable without any skill investment.
  \item[Reactive adaptive.] \textsc{GreedySkillGap} trains toward the largest
    anticipated certified-capacity shortfall (using the forecast lookahead and
    current backlog), committing only the minimum number of workers needed,
    then produces greedily; it is deliberately myopic. \textsc{Balanced\-Heuristic}
    is the same rule with a low training fraction per shift and serves as a
    documented slow-reacting variant.
  \item[Water-filling adaptive.] \textsc{WaterFillingSkillGap} uses the same
    training rule and the same information as \textsc{GreedySkillGap} but
    splits production capacity across products in proportion to need, so no
    product can take the whole pool. It isolates a myopic-allocation artifact
    (Section~\ref{sec:results-wf}) from environment properties.
  \item[Static insurance.] \textsc{StaticTrainingPlan} executes a fixed,
    open-loop cross-training plan in the first five shifts (the two skill-1
    backups and the two rare-skill candidates, 4 hours per worker per shift;
    80 training hours in total) and otherwise produces greedily. It observes
    nothing about shocks.
  \item[Static budget variants.] \textsc{StaticTrainingPlan}$\{40,60,120\}$
    vary \emph{only} the hours per shift of the same plan (2, 3, and 6 hours);
    plan structure and timing are never re-tuned.
\end{description}

\textsc{RandomValid}, an exact-feasible random policy, anchors the bottom of
every table.

\paragraph{Fairness of the static baseline.}
\textsc{StaticTrainingPlan} is a \emph{favorable} static-insurance baseline,
not an unbiased deployment policy: its plan structure encodes exactly which
skills can become critical in this scenario family. It pre-trains the two
contingencies the environment can realize (skill-1 redundancy for absenteeism,
rare-skill coverage for the new product), which a deployed planner would not
know in advance. We therefore interpret it as a strong, structurally informed
comparator that measures the value of insurance, and we report its budget
sensitivity explicitly. In larger instances with more skills or uncertain
skill demand, blanket insurance scales with the number of contingencies while
reactive training scales with realized needs; the small instance studied here
is the static plan's best case.

\subsection{Metrics and statistical protocol}
\label{sec:protocol}

We report operational metrics (total cost, service level, per-product service,
total and peak backlog), resilience metrics (recovery time relative to the
pre-shock backlog level, recovery rate, number of unrecovered seeds),
capability metrics (skill gain, new certifications), training-access metrics
(training hours per worker, minimum access, Jain index, Gini coefficient), and
projection diagnostics (frequency, norm, certification-zeroed hours).

The main suite runs 20 seeds per scenario; the surprise-shock validation and
sensitivity experiments run 50 seeds. Policy comparisons are paired by seed.
The headline statistic is the per-seed win rate on total cost, supported by an
exact two-sided sign test and a seeded paired bootstrap (10{,}000 resamples)
percentile confidence interval on the mean cost difference; a comparison is
called significant only when the interval excludes zero. We report both
statistics because they can disagree when a few large losses skew the mean,
and we flag one borderline interval as seed-sensitive rather than claiming
significance from a single bootstrap draw.

The full suite (four runners, 20--50 seeds per cell) executes in well under a
minute on a laptop; commands and per-seed CSVs accompany the benchmark
specification.

\section{Results}
\label{sec:results}

We organize the results as a regime analysis rather than a single ranking: the
central finding is that no policy class dominates across disruption types, and
that the adaptive-versus-static comparison is governed by bottleneck
visibility, capacity slack, and the forgetting rate.

\subsection{Main benchmark results}
\label{sec:results-main}

Table~\ref{tab:main} reports mean total cost and service level over 20 seeds
for all nine scenarios. Three patterns hold throughout. First,
training-capable policies dominate the production-only baseline in every
scenario: \textsc{GreedyProduction} loses all 20 paired seeds to every
training-capable policy (sign test $p = 1.9\times10^{-6}$), with mean cost
gaps between $14$k and $66$k. Because skills decay, its initial
certifications erode mid-horizon and its service drops to $0.88$ \emph{even
without any shock}; maintenance training is mandatory on long horizons, not
optional (Section~\ref{sec:results-forgetting} isolates this mechanism).
Second, in the new-product scenarios the production-only policy serves none of
the new product's demand (per-product service $0.0$): no worker is initially
certified for the rare skill, so its aggregate service is capped at the
structural ceiling set by the remaining product. Third, the cost ordering
among training-capable policies changes with the disruption type, which the
remainder of this section unpacks.

\begin{table}[t]
\centering
\caption{Main T$=60$ benchmark: mean total cost (mean service level) over 20
seeds. Lower cost is better. RV = \textsc{RandomValid},
GP = \textsc{GreedyProduction}, GSG = \textsc{GreedySkillGap},
BH = \textsc{BalancedHeuristic}, ST = \textsc{StaticTrainingPlan} (80h).}
\label{tab:main}
\small
\begin{tabular}{lrrrrr}
\toprule
Scenario & RV & GP & GSG & BH & ST \\
\midrule
no\_shock            & 211{,}988 (0.48) & 15{,}757 (0.88) & \textbf{1{,}893} (1.00) & 1{,}893 (1.00) & 3{,}362 (1.00) \\
demand early         & 264{,}068 (0.44) & 21{,}233 (0.89) & 8{,}077 (1.00) & \textbf{6{,}958} (1.00) & 9{,}308 (1.00) \\
demand mid           & 246{,}148 (0.44) & 23{,}166 (0.88) & 7{,}577 (1.00) & \textbf{6{,}644} (1.00) & 8{,}244 (1.00) \\
demand late          & 225{,}988 (0.44) & 15{,}872 (0.89) & 7{,}669 (0.99) & \textbf{6{,}627} (0.99) & 7{,}802 (0.99) \\
absenteeism early    & 232{,}415 (0.45) & 24{,}520 (0.88) & 7{,}677 (1.00) & 7{,}252 (1.00) & \textbf{7{,}075} (1.00) \\
absenteeism mid      & 225{,}219 (0.45) & 35{,}180 (0.83) & 7{,}260 (1.00) & 6{,}832 (1.00) & \textbf{6{,}152} (1.00) \\
absenteeism late     & 217{,}363 (0.45) & 17{,}029 (0.87) & 7{,}221 (0.99) & 7{,}143 (0.99) & \textbf{5{,}857} (1.00) \\
new product (ann.)   & 275{,}246 (0.40) & 67{,}925 (0.77) & \textbf{2{,}080} (1.00) & 2{,}102 (1.00) & 2{,}756 (1.00) \\
new product (surp.)  & 278{,}541 (0.38) & 44{,}306 (0.82) & 2{,}880 (0.99) & 4{,}505 (1.00) & \textbf{2{,}630} (1.00) \\
\bottomrule
\end{tabular}
\end{table}

\subsection{Announced versus surprise new-product shock}
\label{sec:results-visibility}

Visibility of the bottleneck is the cleanest separator between policy classes
(Table~\ref{tab:visibility}). When the new product is announced through the
forecast window, forecast-aware adaptive policies pre-train just in time and
beat the 80h static plan on every paired seed (50/50). When the activation is
a surprise with seed-randomized onset, the same comparison is a statistical
tie for the reactive policies (21--29 and 23--27 for \textsc{GreedySkillGap}
and \textsc{WaterFillingSkillGap}; the slow-reacting
\textsc{BalancedHeuristic} loses 1--49). Stratifying the 50-seed comparison by onset
(bins of 14/17/19 seeds) shows the tie is uniform across early, middle, and
late onsets rather than an artifact of where shocks land.

\begin{table}[t]
\centering
\caption{Adaptive vs.\ the 80h static plan on the new-product shock: per-seed
win rate on cost (wins--losses), exact sign-test $p$; 50 seeds throughout.
Onset bins (0--20\,/\,21--35\,/\,36--60) refer to the surprise activation
shift. GSG = \textsc{GreedySkillGap}, WF = \textsc{WaterFillingSkillGap},
BH = \textsc{BalancedHeuristic}.}
\label{tab:visibility}
\small
\begin{tabular}{llll}
\toprule
vs.\ ST(80h) & Announced & Surprise & By onset \\
\midrule
GSG & 1.00 (50--0), $p{<}10^{-14}$ & 0.42 (21--29), $p{=}0.32$ & 7/14 \,/\, 7/17 \,/\, 7/19 \\
WF  & 1.00 (50--0), $p{<}10^{-14}$ & 0.46 (23--27), $p{=}0.67$ & 7/14 \,/\, 9/17 \,/\, 7/19 \\
BH  & 1.00 (50--0), $p{<}10^{-14}$ & 0.02 (1--49), $p{=}9.1{\times}10^{-14}$ & --- \\
\bottomrule
\end{tabular}
\end{table}

\subsection{Isolating the myopic-allocation artifact}
\label{sec:results-wf}

Near capacity, the myopic greedy allocator oscillates: after a surprise
activation, total demand ($\approx 30$h) is close to the pool (32h), and
serving the scarcest product to exhaustion each shift starves the other,
producing a multi-shift backlog ping-pong (worst seeds cost an extra
1{,}200--2{,}300). \textsc{WaterFillingSkillGap} — identical training rule,
identical information, proportional capacity split — removes the oscillation
and beats \textsc{GreedySkillGap} on 45/50 surprise seeds
($p = 4.2\times10^{-9}$, mean $-221$ per episode). Crucially, the thrash-free
allocator still only \emph{ties} the 80h static plan (23--27, n.s., mean
difference $-18.9$ on a total cost of ${\approx}2{,}750$, i.e.\ ${<}1\%$):
the static plan's competitiveness under surprise is genuine insurance
economics, not an artifact of a weak adaptive baseline.

\subsection{Static training-budget sensitivity}
\label{sec:results-budget}

Table~\ref{tab:budget} varies only the static plan's hours per shift
(structure and timing fixed). The 80h calibration is the static plan's
\emph{worst} reasonable setting, not its best: lean plans (40h and 60h) buy
the shock-critical certifications at roughly half the diverted labor and beat
both adaptive policies on 47--49/50 surprise seeds — at the default
forgetting rate \emph{and} with forgetting disabled. Cost sits on a plateau
(${\approx}2{,}020$--$2{,}055$) from 36h to 60h and rises sharply at 80h
(2{,}771) and 120h (5{,}801); 30h fails to certify the rare skill. Precision
matters here: the 40h plan sustains only the shock-critical skill-2
certifications through the horizon, while its skill-1 cross-training decays
back below threshold — half its insurance is wasted in this scenario family,
which is exactly why leaner is cheaper. The visibility result, by contrast,
survives the budget attack: announced-shock adaptive policies beat every
over-provisioned plan decisively and at worst tie the leanest one.

\begin{table}[t]
\centering
\caption{Static budget sensitivity: adaptive-vs-static per-seed win rates on
cost (50 seeds; sign test). ``Surprise $\delta{=}0$'' disables forgetting.
Static mean cost under surprise shown in the last row.}
\label{tab:budget}
\small
\begin{tabular}{lcccc}
\toprule
 & ST(40h) & ST(60h) & ST(80h) & ST(120h) \\
\midrule
WF vs.\ ST, surprise            & 0.02 & 0.02 & 0.46 & 1.00 \\
GSG vs.\ ST, surprise           & 0.04 & 0.04 & 0.42 & 1.00 \\
WF vs.\ ST, surprise $\delta=0$ & 0.06 & 0.06 & 0.84 & 1.00 \\
WF vs.\ ST, announced           & 0.64 & 0.76 & 1.00 & 1.00 \\
GSG vs.\ ST, announced          & 0.66 & 0.92 & 1.00 & 1.00 \\
\midrule
ST mean cost (surprise)         & 2{,}026 & 2{,}053 & 2{,}771 & 5{,}801 \\
\bottomrule
\end{tabular}
\end{table}

\subsection{Capacity slack and the recovery boundary}
\label{sec:results-boundary}

Table~\ref{tab:boundary} reports recovery diagnostics for the water-filling
policy across capacity levels at two demand scales. The boundary is not a
smooth axis: recovery collapses within roughly one labor-hour of the
demand-equals-capacity point and is restored by two units of slack, at both
demand scales — the effect tracks \emph{relative} slack, not absolute
capacity. At zero slack any reaction transient is structurally unrecoverable,
so the static plan dominates regardless of the forgetting rate (at
$\delta{=}0$ the adaptive-vs-ST(80h) win count climbs 5/50 $\to$ 20/50 $\to$
42/50 across capacity 30/31/32). Win rates at the smaller demand scale remain
static-favored at every slack level for a different reason: pre-shock demand
leaves ${\approx}16$h of spare labor, so the static plan's training is nearly
free (mean pre-shock backlog ${\approx}0.3$/step vs.\ ${\approx}3.9$/step at
default demand). The strategic value of insurance is governed by pre-shock
labor utilization; the structural feasibility of reaction is governed by
post-shock slack.

\begin{table}[t]
\centering
\caption{Recovery boundary for \textsc{WaterFillingSkillGap} under the
surprise shock (50 seeds): recovery rate, mean recovery time given recovery,
and unrecovered seeds, by capacity slack relative to post-activation demand
(30h at default scale; 24h at the scaled cells).}
\label{tab:boundary}
\small
\begin{tabular}{llcccc}
\toprule
Demand scale & Slack: & 0 & 1 & 2 & 8 \\
\midrule
default (20+10) & recovery rate & 0.06 & 0.76 & 0.98 & 0.96 \\
                & mean recovery & 32.3 & 15.1 & 8.9  & 9.2 \\
                & unrecovered   & 47   & 12   & 1    & 2 \\
scaled (16+8)   & recovery rate & 0.24 & 0.96 & 0.98 & 1.00 \\
                & mean recovery & 15.3 & 11.0 & 5.9  & 2.7 \\
                & unrecovered   & 38   & 2    & 1    & 0 \\
\bottomrule
\end{tabular}
\end{table}

\subsection{Forgetting sensitivity}
\label{sec:results-forgetting}

Table~\ref{tab:forgetting} disables and scales forgetting in the no-shock
scenario. With $\delta = 0$ or $\delta = 0.0025$, the production-only policy
holds full service: its collapse at the default rate is entirely
forgetting-driven maintenance, cleanly separated from any shock effect. At
$\delta = 0.01$ its service halves while adaptive maintenance training
contains the cost increase to ${\approx}56\%$. Forgetting also conditions the
adaptive-versus-static comparison: disabling it flips the surprise verdict
against the over-provisioned 80h plan (42/50 for water-filling,
$p = 1.2\times10^{-6}$) — blind early fortification has lasting value only
when skills decay — but the flip does not extend to the lean static plans
(Table~\ref{tab:budget}) nor to zero-slack capacity
(Section~\ref{sec:results-boundary}).

\begin{table}[t]
\centering
\caption{Forgetting sensitivity in no\_shock (20 seeds): production-only
service collapses as $\delta$ grows; adaptive maintenance holds service at
bounded cost. GSG cost equals GP cost at $\delta \le 0.0025$ because no
maintenance is needed.}
\label{tab:forgetting}
\small
\begin{tabular}{lcccc}
\toprule
$\delta$ (per shift) & 0 & 0.0025 & 0.005 & 0.01 \\
\midrule
GP service           & 1.00 & 1.00 & 0.88 & 0.54 \\
GP mean cost         & 1{,}924 & 1{,}924 & 15{,}757 & 96{,}542 \\
GSG mean cost        & 1{,}924 & 1{,}924 & 1{,}893 & 2{,}949 \\
ST(80h) mean cost    & --- & --- & 3{,}362 & --- \\
\bottomrule
\end{tabular}
\end{table}

\begin{figure}[t]
\centering
\includegraphics[width=\linewidth]{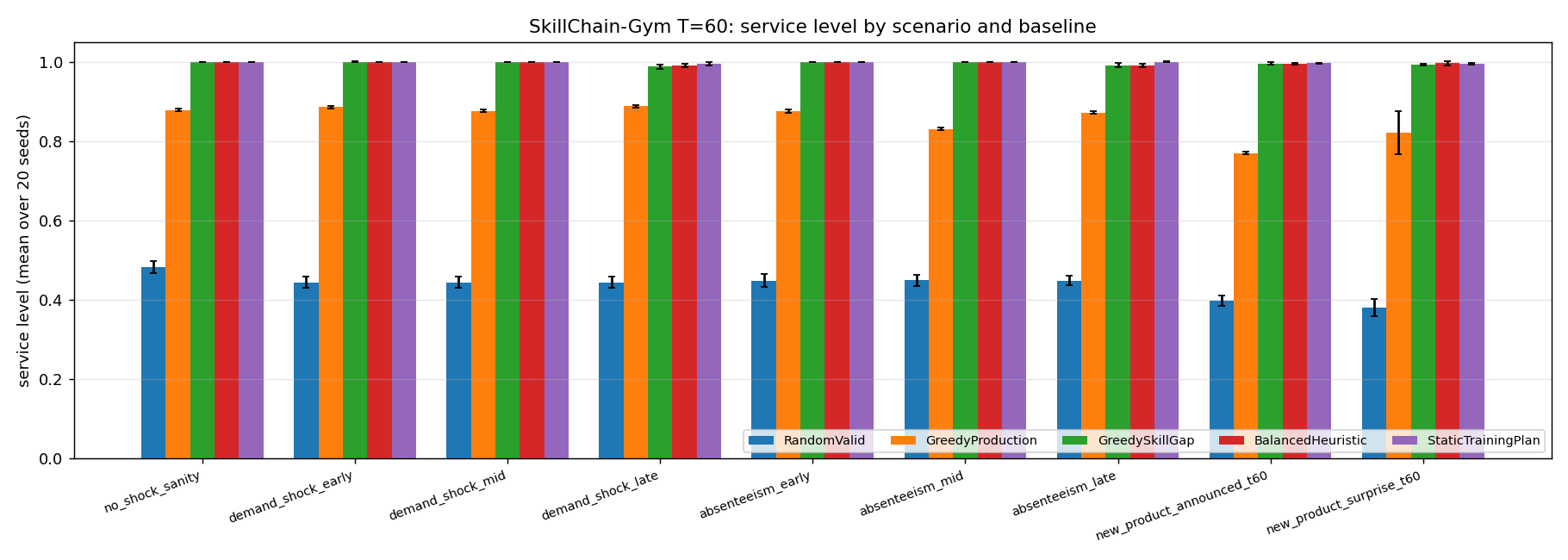}
\caption{Service level by scenario and baseline (mean $\pm$ s.d.\ over 20
seeds, T$=60$). Training-capable policies hold near-full service everywhere;
the production-only baseline degrades even without shocks due to forgetting.}
\label{fig:service}
\end{figure}

Figure~\ref{fig:service} summarizes service levels across all scenarios and
baselines; backlog trajectories and the cost-versus-training-hours trade-off
are shown in the supplementary figures.

\section{Discussion and Limitations}
\label{sec:discussion}

\paragraph{What the benchmark shows.}
The experiments support a regime map rather than a winner. Training-capable
policies dominate the production-only baseline under the default benchmark
with forgetting and skill shocks; with realistic skill decay, maintenance
training is mandatory even in the absence of disruptions. Adaptive training
wins when bottlenecks are visible or forecasted: on the announced new-product
shock it beats every over-provisioned static plan decisively and is at worst
statistically indistinguishable from the leanest fully-certifying plan. Under
surprise shocks and absenteeism, a lean static cross-training plan is the
strongest baseline we tested: it acts as insurance whose strategic cost is the
pre-shock labor it diverts and whose value rises near the demand--capacity
boundary, where reaction transients become structurally unrecoverable. This
echoes classical results on cross-training and workforce flexibility as a
buffer against variability
\cite{Hopp2004SkillChaining,Park1991CrossTraining,Valeva2017FlexInventory},
and emerges here without tuning. No policy class dominates across regimes.

\paragraph{Interpreting the static baseline.}
The static plan's strength must be read with its construction in mind: its
plan structure encodes which skills can become critical in this scenario
family, so it pre-buys exactly the right contingencies. It is a strong,
structurally informed comparator --- useful precisely because it bounds what
insurance can achieve --- but it is not an unbiased deployment policy. Its
budget sensitivity reinforces this reading: the leanest budgets win because
less of the (partially wasted) insurance competes with production, and in
larger instances with more skills or uncertain skill demand, blanket insurance
scales with the number of contingencies while reactive training scales with
realized needs. A scenario family rich enough that no static plan can
pre-train every contingency is deliberately left as future work and as the
natural bridge to forecast-driven control.

\paragraph{Baseline quality matters.}
The water-filling comparison shows that part of what looks like a policy-class
difference can be an allocation artifact: a myopic greedy allocator thrashes
between products near capacity, and replacing only the allocation rule (same
training, same information) recovers most of that loss. Benchmark conclusions
about ``adaptive vs.\ static'' should therefore always be read jointly with
the allocator used; we report both the myopic and the water-filling variants
for this reason.

\paragraph{Mechanisms, not just rankings.}
Two mechanisms organize the results. First, forgetting converts training from
a one-off investment into a maintenance flow; it explains the production-only
collapse without shocks, and it is the reason blind early fortification
retains value (disabling forgetting flips the comparison against the
over-provisioned static plan, but not against lean plans, and not at zero
slack). Second, capacity slack acts as a boundary, not a smooth axis: recovery
from a surprise collapses within roughly one labor-hour of the
demand-equals-capacity point and is restored by two, at both demand scales
tested. Reporting recovery rates and unrecovered-seed counts alongside cost
means is essential; cost alone would misattribute a structural infeasibility
to a strategic advantage.

\paragraph{Limitations.}
The benchmark is synthetic and stylized by design. Skill dynamics are a
two-parameter abstraction (linear training gain, exponential decay) with hard
threshold certification; no claim is made that they are calibrated to real
workforce behavior. The instance is small (two products, three skills, four
workers, one site), which is the static plan's best case and keeps every
policy interpretable; procurement, supplier delays, multi-echelon flows,
job-shop routing, soft productivity, and learning-by-doing are deliberately
excluded from the default model. The baselines are heuristics; rolling-horizon
optimization and learned policies are deferred, and closed-loop
forecast-driven control is the subject of the companion controller study.
Statistically, sign tests at 20--50 seeds have limited power for small
effects, and one borderline bootstrap interval (the reactive policy against
the 80h plan without forgetting) is seed-sensitive; we therefore lead with
per-seed win rates and report effect sizes for ties (the headline tie is
${<}1\%$ of total cost). Demand noise is a lower-censored Gaussian whose bias
is negligible at the configured noise scale but would matter in low-mean
sweeps; the total labor envelope is a second binding resource that any
scenario scaling must respect. Finally, training-access metrics (per-worker
hours, minimum access, Jain and Gini indices) are reported as observables;
no baseline optimizes them, and no demographic fairness claims are made.

\paragraph{Outlook.}
The regime structure identified here --- visibility, slack, and forgetting as
the governing parameters --- is exactly the setting in which receding-horizon
controllers that value future skill capacity should help: they can buy
insurance when forecasts warrant it and react efficiently when they do not.
The companion paper studies such skill-constrained controllers on this
benchmark.

\section{Conclusion}
\label{sec:conclusion}

We presented SkillChain-Gym, a benchmark specification for reskilling-aware
production-inventory control at a single site. The benchmark joins two
research lines that have largely developed separately: reusable
operations-research environments, which treat labor as exogenous, and
workforce-planning models with skills, learning, and forgetting, which are
rarely released as shared testbeds. Its locked formulation makes worker
capability part of the state --- continuous skill levels with hard threshold
certification and forgetting --- and makes reskilling part of the action
space, as a capacity-consuming allocation that competes with production for
the same worker hours. Seed-controlled disruption scenarios, three
feasibility modes with mandatory projection diagnostics, deterministic
replay, and metrics spanning operations, resilience, capability growth, and
training-access distribution complete the specification, together with an
exact-feasible baseline taxonomy: production-only, reactive adaptive,
water-filling adaptive, and static-insurance policies with budget variants.

The empirical message is a regime map, not a ranking. Training-capable
policies dominate the production-only baseline in every scenario; under
realistic forgetting, maintenance training is mandatory even without shocks.
Between training-capable classes, however, no policy dominates: adaptive
training wins when bottlenecks are visible or forecasted, while a lean static
cross-training plan --- a deliberately favorable comparator whose structure
encodes which skills can become critical --- acts as strong insurance under
surprise shocks and absenteeism, with capacity slack and the forgetting rate
governing where each class prevails. The benchmark's value lies in making
these regimes measurable and reproducible rather than in crowning a method.

The regime structure also defines the next step. A controller that uses
forecasts to decide \emph{when} to buy insurance and \emph{when} to react ---
valuing future skill capacity explicitly over a receding horizon --- should
be able to combine the strengths of both policy classes; we study such
skill-constrained, forecast-driven control on this benchmark in a companion
paper. Richer scenario families, including instances with enough skills that
no static plan can pre-train every contingency, and calibration of demand
patterns and skill taxonomies to public data sources, are natural extensions
of the specification.

\section*{Funding}
No external funding was received for this work.

\bibliographystyle{plain}
\bibliography{references}

@misc{Alvo2023InventoryNetworks,
  title = {Deep Reinforcement Learning for Inventory Networks: Toward Reliable Policy Optimization},
  author = {Alvo, Matias and Russo, Daniel and Kanoria, Yash and Lee, Minuk},
  year = {2023},
  eprint = {2306.11246},
  archivePrefix = {arXiv},
  primaryClass = {cs.LG},
  doi = {10.48550/arXiv.2306.11246},
  url = {https://arxiv.org/abs/2306.11246}
}

@article{Azizi2010JobRotation,
  title = {Modeling job rotation in manufacturing systems: The study of employee's boredom and skill variations},
  author = {Azizi, Nader and Zolfaghari, Saeed and Liang, Ming},
  journal = {International Journal of Production Economics},
  volume = {123},
  number = {1},
  pages = {69--85},
  year = {2010},
  doi = {10.1016/j.ijpe.2009.07.010},
  url = {https://doi.org/10.1016/j.ijpe.2009.07.010}
}

@inproceedings{Berto2025RL4CO,
  title = {{RL4CO}: An Extensive Reinforcement Learning for Combinatorial Optimization Benchmark},
  author = {Berto, Federico and Hua, Chuanbo and Park, Junyoung and Luttmann, Laurin and Ma, Yining and Bu, Fanchen and Wang, Jiarui and Ye, Haoran and Kim, Minsu and Choi, Sanghyeok and Gast Zepeda, Nayeli and Hottung, Andr{\'e} and Zhou, Jianan and Bi, Jieyi and Hu, Yu and Liu, Fei and Kim, Hyeonah and Son, Jiwoo and Kim, Haeyeon and Angioni, Davide and Kool, Wouter and Cao, Zhiguang and Zhang, Qingfu and Kim, Joungho and Zhang, Jie and Shin, Kijung and Wu, Cathy and Ahn, Sungsoo and Song, Guojie and Kwon, Changhyun and Tierney, Kevin and Xie, Lin and Park, Jinkyoo},
  booktitle = {Proceedings of the 31st ACM SIGKDD Conference on Knowledge Discovery and Data Mining V.2},
  pages = {5278--5289},
  year = {2025},
  doi = {10.1145/3711896.3737433},
  url = {https://doi.org/10.1145/3711896.3737433}
}

@article{Burtea2024ConstrainedInventoryRL,
  title = {Constrained continuous-action reinforcement learning for supply chain inventory management},
  author = {Burtea, Radu and Tsay, Calvin},
  journal = {Computers \& Chemical Engineering},
  volume = {181},
  pages = {108518},
  year = {2024},
  doi = {10.1016/j.compchemeng.2023.108518},
  url = {https://doi.org/10.1016/j.compchemeng.2023.108518}
}

@article{Cavagnini2020UncertainLearning,
  title = {Workforce production planning under uncertain learning rates},
  author = {Cavagnini, Rossana and Hewitt, Mike and Maggioni, Francesca},
  journal = {International Journal of Production Economics},
  volume = {225},
  pages = {107590},
  year = {2020},
  doi = {10.1016/j.ijpe.2019.107590},
  url = {https://doi.org/10.1016/j.ijpe.2019.107590}
}

@article{DeBruecker2015SkillsReview,
  title = {Workforce planning incorporating skills: State of the art},
  author = {De Bruecker, Philippe and Van den Bergh, Jorne and Beli{\"e}n, Jeroen and Demeulemeester, Erik},
  journal = {European Journal of Operational Research},
  volume = {243},
  number = {1},
  pages = {1--16},
  year = {2015},
  doi = {10.1016/j.ejor.2014.10.038},
  url = {https://doi.org/10.1016/j.ejor.2014.10.038}
}

@article{Felan2001HeterogeneousFlexibility,
  title = {Multi-level heterogeneous worker flexibility in a Dual Resource Constrained (DRC) job-shop},
  author = {Felan, Joe T. and Fry, Timothy D.},
  journal = {International Journal of Production Research},
  volume = {39},
  number = {14},
  pages = {3041--3059},
  year = {2001},
  doi = {10.1080/00207540110047702},
  url = {https://doi.org/10.1080/00207540110047702}
}

@article{Geevers2024MultiEchelonDRL,
  title = {Multi-echelon inventory optimization using deep reinforcement learning},
  author = {Geevers, Kevin and van Hezewijk, Lotte and Mes, Martijn R. K.},
  journal = {Central European Journal of Operations Research},
  volume = {32},
  number = {3},
  pages = {653--683},
  year = {2024},
  doi = {10.1007/s10100-023-00872-2},
  url = {https://doi.org/10.1007/s10100-023-00872-2}
}

@article{Grumbach2023Sociotechnical,
  title = {A Memetic Algorithm With Reinforcement Learning for Sociotechnical Production Scheduling},
  author = {Grumbach, Felix and Badr, Nour Eldin Alaa and Reusch, Pascal and Trojahn, Sebastian},
  journal = {IEEE Access},
  volume = {11},
  pages = {68760--68775},
  year = {2023},
  doi = {10.1109/ACCESS.2023.3292548},
  url = {https://doi.org/10.1109/ACCESS.2023.3292548}
}

@article{Heide2026AIAssistance,
  title = {AI-driven collaborative assistance system in production planning},
  author = {Heide, Klaas Maximilian and Wichmann, Marcel and Settnik, Simon Johannes and Voelker, Hendrik and M{\"u}nch, Gina Vibora},
  journal = {CIRP Journal of Manufacturing Science and Technology},
  volume = {67},
  pages = {48--59},
  year = {2026},
  doi = {10.1016/j.cirpj.2026.02.010},
  url = {https://doi.org/10.1016/j.cirpj.2026.02.010}
}

@article{Heuser2022BudgetedTraining,
  title = {Workforce planning in production with flexible or budgeted employee training and volatile demand},
  author = {Heuser, Patricia and Letmathe, Peter and Schinner, Matthias},
  journal = {Journal of Business Economics},
  volume = {92},
  number = {7},
  pages = {1093--1124},
  year = {2022},
  doi = {10.1007/s11573-022-01090-z},
  url = {https://doi.org/10.1007/s11573-022-01090-z}
}

@article{Heuser2023SingleMachineLearningForgetting,
  title = {Single-machine scheduling with product category-based learning and forgetting effects},
  author = {Heuser, Patricia and Tauer, Bj{\"o}rn},
  journal = {Omega},
  volume = {115},
  pages = {102786},
  year = {2023},
  doi = {10.1016/j.omega.2022.102786},
  url = {https://doi.org/10.1016/j.omega.2022.102786}
}

@article{Hopp2004SkillChaining,
  title = {Benefits of Skill Chaining in Serial Production Lines with Cross-Trained Workers},
  author = {Hopp, Wallace J. and Tekin, Eylem and Van Oyen, Mark P.},
  journal = {Management Science},
  volume = {50},
  number = {1},
  pages = {83--98},
  year = {2004},
  doi = {10.1287/mnsc.1030.0166},
  url = {https://doi.org/10.1287/mnsc.1030.0166}
}

@article{Hu2025GraphDRLWorkerFatigue,
  title = {Graph-based deep reinforcement learning for dynamic scheduling of flexible job-shop considering worker fatigue and multi-skill factors},
  author = {Hu, Yiwen and Zhang, Zequn and Chen, Jie and Tang, Dunbing and Cai, Qixiang},
  journal = {Applied Soft Computing},
  volume = {184},
  pages = {113712},
  year = {2025},
  doi = {10.1016/j.asoc.2025.113712},
  url = {https://doi.org/10.1016/j.asoc.2025.113712}
}

@misc{Hubbs2020ORGym,
  title = {{OR-Gym}: A Reinforcement Learning Library for Operations Research Problems},
  author = {Hubbs, Christian D. and Perez, Hector D. and Sarwar, Owais and Sahinidis, Nikolaos V. and Grossmann, Ignacio E. and Wassick, John M.},
  year = {2020},
  eprint = {2008.06319},
  archivePrefix = {arXiv},
  primaryClass = {cs.AI},
  doi = {10.48550/arXiv.2008.06319},
  url = {https://arxiv.org/abs/2008.06319}
}

@inproceedings{Jabbari2017FairnessRL,
  title = {Fairness in Reinforcement Learning},
  author = {Jabbari, Shahin and Joseph, Matthew and Kearns, Michael and Morgenstern, Jamie and Roth, Aaron},
  booktitle = {Proceedings of the 34th International Conference on Machine Learning},
  series = {Proceedings of Machine Learning Research},
  volume = {70},
  pages = {1617--1626},
  year = {2017},
  publisher = {PMLR},
  url = {https://proceedings.mlr.press/v70/jabbari17a.html},
  note = {Also available as arXiv:1611.03071}
}

@techreport{Jain1984FairnessIndex,
  title = {A Quantitative Measure Of Fairness And Discrimination For Resource Allocation In Shared Computer Systems},
  author = {Jain, Raj and Chiu, Dah-Ming and Hawe, William},
  institution = {Digital Equipment Corporation},
  number = {DEC Research Report TR-301},
  year = {1984},
  url = {https://www.cs.wustl.edu/~jain/papers/fairness.htm},
  note = {DOI not found}
}

@article{Karimi2024TechnicianResource,
  title = {Robust Optimization for Technician and Resource Management in Reconfigurable Assembly Lines},
  author = {Karimi, Tourandokht and Thevenin, Simon and Benderbal, Hichem Haddou},
  journal = {IFAC-PapersOnLine},
  volume = {58},
  number = {19},
  pages = {295--300},
  year = {2024},
  doi = {10.1016/j.ifacol.2024.09.197},
  url = {https://doi.org/10.1016/j.ifacol.2024.09.197}
}

@article{Kher1994WorkerFlexibility,
  title = {Acquiring and operationalizing worker flexibility in dual resource constrained job shops with worker transfer delays and learning losses},
  author = {Kher, H. V. and Malhotra, M. K.},
  journal = {Omega},
  volume = {22},
  number = {5},
  pages = {521--533},
  year = {1994},
  doi = {10.1016/0305-0483(94)90032-9},
  url = {https://doi.org/10.1016/0305-0483(94)90032-9}
}

@article{Liu2016WorkerAssignment,
  title = {Worker assignment and production planning with learning and forgetting in manufacturing cells by hybrid bacteria foraging algorithm},
  author = {Liu, Chunfeng and Wang, Jufeng and Leung, Joseph Y.-T.},
  journal = {Computers \& Industrial Engineering},
  volume = {96},
  pages = {162--179},
  year = {2016},
  doi = {10.1016/j.cie.2016.03.020},
  url = {https://doi.org/10.1016/j.cie.2016.03.020}
}

@article{Lu2025DisruptiveInventoryPPO,
  title = {Dynamic Optimization of Multi-Echelon Supply Chain Inventory Policies Under Disruptive Scenarios: A Deep Reinforcement Learning Approach},
  author = {Lu, Xiaonong and Wang, Hongzhe and Peng, Zhanglin and Liao, Chen and Liu, Chunyan},
  journal = {Symmetry},
  volume = {17},
  number = {12},
  pages = {2078},
  year = {2025},
  doi = {10.3390/sym17122078},
  url = {https://doi.org/10.3390/sym17122078}
}

@incollection{Palao2023CompetenceAware,
  title = {Framework for Formulating Competence-Aware Scheduling Models in Mixed-Model Assembly},
  author = {Palao, Carlos Miguel and Hoedt, Steven and Leyman, Pieter and Aghezzaf, El-Houssaine and Cottyn, Johannes},
  booktitle = {Production Processes and Product Evolution in the Age of Disruption},
  series = {Lecture Notes in Mechanical Engineering},
  pages = {552--561},
  year = {2023},
  doi = {10.1007/978-3-031-34821-1_60},
  url = {https://doi.org/10.1007/978-3-031-34821-1_60}
}

@incollection{Palao2025ResilientWorkforce,
  title = {Resilient Workforce Planning for Mixed-Model Assembly Lines with Training and Learning Effects},
  author = {Palao, Carlos Miguel and Hoedt, Steven and Aghezzaf, El-Houssaine and Cottyn, Johannes},
  booktitle = {Advances in Production Management Systems. Cyber-Physical-Human Production Systems: Human-AI Collaboration and Beyond},
  series = {IFIP Advances in Information and Communication Technology},
  volume = {769},
  pages = {232--246},
  year = {2025},
  doi = {10.1007/978-3-032-03550-9_16},
  url = {https://doi.org/10.1007/978-3-032-03550-9_16}
}

@article{Park1991CrossTraining,
  title = {The examination of worker cross-training in a dual resource constrained job shop},
  author = {Park, Paul Sungchil},
  journal = {European Journal of Operational Research},
  volume = {52},
  number = {3},
  pages = {291--299},
  year = {1991},
  doi = {10.1016/0377-2217(91)90164-Q},
  url = {https://doi.org/10.1016/0377-2217(91)90164-Q}
}

@article{Preil2022BanditInventory,
  title = {Bandit-based inventory optimisation: Reinforcement learning in multi-echelon supply chains},
  author = {Preil, Deniz and Krapp, Michael},
  journal = {International Journal of Production Economics},
  volume = {252},
  pages = {108578},
  year = {2022},
  doi = {10.1016/j.ijpe.2022.108578},
  url = {https://doi.org/10.1016/j.ijpe.2022.108578}
}

@misc{Ramanujam2025SafeORGym,
  title = {{SafeOR-Gym}: A Benchmark Suite for Safe Reinforcement Learning Algorithms on Practical Operations Research Problems},
  author = {Ramanujam, Asha and Elyoumi, Adam and Chen, Hao and Kompalli, Sai Madhukiran and Ahluwalia, Akshdeep Singh and Pal, Shraman and Papageorgiou, Dimitri J. and Li, Can},
  year = {2025},
  eprint = {2506.02255},
  archivePrefix = {arXiv},
  primaryClass = {cs.LG},
  doi = {10.48550/arXiv.2506.02255},
  url = {https://arxiv.org/abs/2506.02255}
}

@article{Rolf2022SCMRLReview,
  title = {A review on reinforcement learning algorithms and applications in supply chain management},
  author = {Rolf, Benjamin and Jackson, Ilya and M{\"u}ller, Marcel and Lang, Sebastian and Reggelin, Tobias and Ivanov, Dmitry},
  journal = {International Journal of Production Research},
  volume = {61},
  number = {20},
  pages = {7151--7179},
  year = {2022},
  doi = {10.1080/00207543.2022.2140221},
  url = {https://doi.org/10.1080/00207543.2022.2140221}
}

@article{Ruf2022HierarchicalSkills,
  title = {Workforce capacity planning with hierarchical skills, long-term training, and random resignations},
  author = {Ruf, Christian and Bard, Jonathan F. and Kolisch, Rainer},
  journal = {International Journal of Production Research},
  volume = {60},
  number = {2},
  pages = {783--807},
  year = {2022},
  doi = {10.1080/00207543.2021.2017058},
  url = {https://doi.org/10.1080/00207543.2021.2017058}
}

@article{SaidiMehrabad2013WorkerTraining,
  title = {Production planning and worker training in dynamic manufacturing systems},
  author = {Saidi-Mehrabad, Mohammad and Paydar, Mohammad Mahdi and Aalaei, Amin},
  journal = {Journal of Manufacturing Systems},
  volume = {32},
  number = {2},
  pages = {308--314},
  year = {2013},
  doi = {10.1016/j.jmsy.2012.12.007},
  url = {https://doi.org/10.1016/j.jmsy.2012.12.007}
}

@article{Valeva2017FlexInventory,
  title = {Balancing flexibility and inventory in workforce planning with learning},
  author = {Valeva, Silviya and Hewitt, Mike and Thomas, Barrett W. and Brown, Kenneth G.},
  journal = {International Journal of Production Economics},
  volume = {183},
  pages = {194--207},
  year = {2017},
  doi = {10.1016/j.ijpe.2016.10.026},
  url = {https://doi.org/10.1016/j.ijpe.2016.10.026}
}

@article{Waubert2022ReliabilitySurvey,
  title = {On reliability of reinforcement learning based production scheduling systems: a comparative survey},
  author = {Waubert de Puiseau, Constantin and Meyes, Richard and Meisen, Tobias},
  journal = {Journal of Intelligent Manufacturing},
  volume = {33},
  number = {4},
  pages = {911--927},
  year = {2022},
  doi = {10.1007/s10845-022-01915-2},
  url = {https://doi.org/10.1007/s10845-022-01915-2}
}

@misc{Weng2019FairnessRL,
  title = {Fairness in Reinforcement Learning},
  author = {Weng, Paul},
  year = {2019},
  eprint = {1907.10323},
  archivePrefix = {arXiv},
  primaryClass = {cs.LG},
  doi = {10.48550/arXiv.1907.10323},
  url = {https://arxiv.org/abs/1907.10323},
  note = {Presented at the AI for Social Good Workshop at IJCAI 2019}
}

@article{Wu2023DistributionalInventory,
  title = {Distributional reinforcement learning for inventory management in multi-echelon supply chains},
  author = {Wu, Guoquan and de Carvalho Servia, Miguel {\'A}ngel and Mowbray, Max},
  journal = {Digital Chemical Engineering},
  volume = {6},
  pages = {100073},
  year = {2023},
  doi = {10.1016/j.dche.2022.100073},
  url = {https://doi.org/10.1016/j.dche.2022.100073}
}

@incollection{Wu2023InventoryManagementCAChE,
  title = {Reinforcement Learning for inventory management in multi-echelon supply chains},
  author = {Wu, Guoquan and de Carvalho Servia, Miguel {\'A}ngel and Mowbray, Max},
  booktitle = {Computer Aided Chemical Engineering},
  pages = {795--800},
  year = {2023},
  doi = {10.1016/B978-0-443-15274-0.50127-X},
  url = {https://doi.org/10.1016/B978-0-443-15274-0.50127-X}
}

@misc{Yang2023MABIM,
  title = {A Versatile Multi-Agent Reinforcement Learning Benchmark for Inventory Management},
  author = {Yang, Xianliang and Liu, Zhihao and Jiang, Wei and Zhang, Chuheng and Zhao, Li and Song, Lei and Bian, Jiang},
  year = {2023},
  eprint = {2306.07542},
  archivePrefix = {arXiv},
  primaryClass = {cs.AI},
  doi = {10.48550/arXiv.2306.07542},
  url = {https://arxiv.org/abs/2306.07542}
}

@article{Yilmaz2017FairSharing,
  title = {A framework and algorithm for fair demand and capacity sharing in collaborative networks},
  author = {Yilmaz, Ibrahim and Yoon, Sang Won and Seok, Hyesung},
  journal = {International Journal of Production Economics},
  volume = {193},
  pages = {137--147},
  year = {2017},
  doi = {10.1016/j.ijpe.2017.06.027},
  url = {https://doi.org/10.1016/j.ijpe.2017.06.027}
}

@article{Yue2008WorkerFlexibility,
  title = {Worker flexibility in a parallel dual resource constrained job shop},
  author = {Yue, H. and Slomp, J. and Molleman, E. and van der Zee, D. J.},
  journal = {International Journal of Production Research},
  volume = {46},
  number = {2},
  pages = {451--467},
  year = {2008},
  doi = {10.1080/00207540601138510},
  url = {https://doi.org/10.1080/00207540601138510}
}

@misc{wef_reskilling_2026,
  title = {Reskilling Revolution: Preparing 1 billion people for tomorrow's economy},
  author = {{World Economic Forum}},
  year = {2026},
  howpublished = {\url{https://www.weforum.org/impact/reskilling-revolution-preparing-1-billion-people-for-tomorrows-economy/}},
  note = {Accessed 2026-06-06}
}

\end{document}